\documentclass{article}
\pdfoutput=1

\usepackage{times}
\usepackage{url}
\usepackage{natbib}
\usepackage[usenames,dvipsnames]{color}
\usepackage{xcolor}
\usepackage{float}
\usepackage{graphicx}
\usepackage{hyperref}
\usepackage{latexsym}
\usepackage{amssymb, amsmath}
\usepackage{multirow}
\usepackage{subcaption}
\usepackage{listings}
\usepackage[normalem]{ulem}
\usepackage[font=normalsize, labelfont=bf, margin=6pt]{caption}
\usepackage[top=1in, bottom=1in, left=1.5in, right=1.5in,headheight=110pt]{geometry}

\definecolor{lightgray}{gray}{0.7}
\definecolor{anti-flashwhite}{rgb}{0.95, 0.95, 0.96}
\definecolor{coolgrey}{rgb}{0.55, 0.57, 0.67}
\definecolor{arsenic}{rgb}{0.23, 0.27, 0.29}

\newcommand{\capcode}[1]{{\tt {\footnotesize #1}}}

\newcommand{\lineref}[1]{\hyperref[#1]{\textsc{line} \ref{#1}}}

\lstdefinelanguage{scala}{
 morekeywords={abstract,case,catch,class,def,%
  do,else,extends,false,final,finally,%
  for,if,implicit,import,match,mixin,%
  new,null,object,override,package,%
  private,protected,requires,return,sealed,%
  super,this,throw,trait,true,try,%
  type,val,var,while,with,yield,Nil,%
  _,=>,<-,<\%,<:,>:,\#,@},
 sensitive=true,
 morecomment=[l]{//},
 morecomment=[n]{/*}{*/},
 commentstyle=\color{coolgrey}\textit,
 morestring=[b]",
 morestring=[b]',
 morestring=[b]"""
}

\lstdefinelanguage{yaml}{
  morestring=[b]", keywords={true, false, null, name, action, example,
    priority, type, label, unit, pattern}, keywordstyle=\textbf,
  ndkeywords={trigger, ?<trigger>, theme, theme?, ?<theme>, @theme,
    cause, cause?, ?<cause>, @cause, site, site?, @site, product,
    product?, @site, controller, controlled, @controlled, 
    ?<witness>, @witness, 
    ?<entity>, @entity, entity, 
    ?<location>, @location, location,
    ?<robber>, @robber, robber,
    @property, property,
    agent, vehicle,
    obstacle,
    title,
    @person, person,
    @resident, resident,
    @num,
    dancer, partner,
    date, spouse,
    city},
  ndkeywordstyle=\color{arsenic}\bfseries, comment=[l]{\#},
  commentstyle=\color{coolgrey}\textit, stringstyle=\ttfamily,
  sensitive=true }

\lstdefinestyle{scala-style}{%
  language=scala,
  basicstyle=\ttfamily\footnotesize,
  basewidth={0.5em,0.5em},
  xleftmargin=9pt,
  xrightmargin=-8pt,
  framexleftmargin=12pt,
  framexrightmargin=-10pt,
  frame=tb,
  captionpos=t,
  rulecolor=\color{black},
  extendedchars=true,
  showstringspaces=false,
  showspaces=false,
  numbers=left,
  numberstyle=\tiny,
  numbersep=8pt,
  stepnumber=1,
  tabsize=2,
  breaklines=true,
  showtabs=false,
  escapeinside={(*@}{@*)}
}

\lstdefinestyle{yaml-style}{%
  language=yaml,
  basicstyle=\ttfamily\footnotesize,
  xleftmargin=9pt,
  xrightmargin=-8pt,
  framexleftmargin=12pt,
  framexrightmargin=-10pt,
  frame=tb,
  captionpos=t,
  rulecolor=\color{black},
  extendedchars=true,
  showstringspaces=false,
  showspaces=false,
  numbers=left,
  numberstyle=\tiny,
  numbersep=8pt,
  stepnumber=1,
  tabsize=2,
  breaklines=true,
  showtabs=false,
  escapeinside={(*@}{@*)}
}

\lstnewenvironment{scala}[1]{%
 \lstset{style=scala-style}}{%
\captionof{lstlisting}{#1}
}

\lstnewenvironment{yaml}[1]{%
 \lstset{style=yaml-style}}{%
\captionof{lstlisting}{#1}
}

\hypersetup{pdftitle={Description of the Odin Event Extraction Framework and Rule Language}}

\title{Description of the Odin Event Extraction Framework and Rule Language}

\author{Marco A. Valenzuela-Esc\'{a}rcega\quad
Gus Hahn-Powell\quad
Mihai Surdeanu \\ 
Computational Language Understanding (CLU) Lab \\
University of Arizona, Tucson, AZ, USA \\
{\tt \{marcov,hahnpowell,msurdeanu\}@email.arizona.edu} \\
}
\date{Last Revised: \today\\
Version 1.0 (see Changes)
}
\begin{document}
\maketitle

\begin{abstract}
This document describes the Odin framework, which is a domain-independent platform for developing rule-based event extraction models. Odin aims to be powerful (the rule language allows the modeling of complex syntactic structures) and robust (to recover from syntactic parsing errors, syntactic patterns can be freely mixed with surface, token-based patterns), while remaining simple (some domain grammars can be up and running in minutes), and fast (Odin processes over 100 sentences/second in a real-world domain with over 200 rules). Here we include a thorough definition of the Odin rule language, together with a description of the Odin API in the Scala language, which allows one to apply these rules to arbitrary texts. 
\end{abstract}

\newpage
\tableofcontents

\newpage
\section{Changes}
\begin{itemize}
\item[1.0:] Initial release.
\end{itemize}

\section{Introduction}
Rule-based information extraction (IE) has long enjoyed wide adoption throughout industry, though it has remained largely ignored in academia, in favor of machine learning (ML) methods \citep{chiticariu2013rule}. However, rule-based systems have several advantages over pure ML systems, including: (a) the rules are interpretable and thus suitable for rapid development and domain transfer; and (b) humans and machines can contribute to the same model. Why then have such systems failed to hold the attention of the academic community? One argument raised by Chiticariu et al. is that, despite notable efforts~\citep{appelt1998common, levy06, hunter2008opendmap, Cunningham11, tokensregex2014}, there is not a standard language for this task, or a ``standard way to express rules'', 
which raises the entry cost for new rule-based systems. 

Odin (Open Domain INformer) aims to address this issue with a novel event extraction (EE) language and framework.
The design of Odin followed the simplicity principles promoted by other natural language processing toolkits, such as Stanford's CoreNLP, which aim to ``avoid over-design'', ``do one thing well'', and have a user ``up and running in ten minutes or less''~\citep{manning14}. For example, consider a domain that tracks people's movement, as reported in the news. One may want to quickly write a domain grammar that captures events with the following arguments: (a) the subject of the verb ``move'' (and its synonyms) only if it has been identified as a \textsc{person} by a named entity recognizer (NER), (b) the indirect object of the same verb that is dominated by the preposition ``from'' as the origin \textit{location}, and (c) an indirect object dominated by the preposition ``to'' as the \textit{destination}. Odin captures such event patterns (and more) using a single declarative rule.

In particular, Odin is:

{\flushleft {\bf Simple:}} Taking advantage of a syntactic dependency (SD) representation~\citep{Marneffe:08}, our EE language has a simple, declarative syntax for the extraction of \textit{n}-ary events, which captures single or multi-word event predicates with lexical and morphological constraints, and event arguments with (generally) simple syntactic patterns and semantic constraints.

{\flushleft {\bf Powerful:}} Despite its simplicity, our EE framework can capture complex constructs when necessary, such as: (a) recursive events\footnote{Events that take other events as arguments. See the walkthrough example in the next section.}, and (b) complex regular expressions over syntactic patterns for event arguments. Inspired by Stanford's Semgrex\footnote{\url{nlp.stanford.edu/software/tregex.shtml}}, we have extended a standard regular expression language to describe patterns over directed graphs\footnote{We currently use Stanford syntactic dependencies, but any other graph derived from text could be used.  For example, one could use a graph that models semantic roles or abstract meaning representation.}, e.g., we introduce new \capcode{$<$} and \capcode{$>$} operators to specify the direction of edge traversal in the dependency graph. Finally, we allow for (c) optional arguments and multiple arguments with the same name.

{\flushleft {\bf Robust:}} To recover from unavoidable syntactic errors, SD patterns (such as the ones shown in the next section) can be can be freely mixed with token-based surface patterns, using a language inspired by the Allen Institute of Artificial Intelligence's Tagger\footnote{\url{https://github.com/allenai/taggers}}. These patterns match against information extracted in our text processing pipeline\footnote{\url{https://github.com/sistanlp/processors}}
, namely a token's part of speech, lemmatized form, named entity label, and the immediate incoming and outgoing edges in the SD graph.  

{\flushleft {\bf Fast:}} 
 Our EE runtime is fast because the Odin runtime uses event trigger phrases (e.g., ``move'' for a moving event), which are captured with lexico-morphological patterns, as shallow filters to reduce the search space for pattern matching. That is, only when event triggers are detected is the matching of more complex syntactic patterns for arguments attempted. This guarantees quick executions. 
 For example, in a real-world biochemical domain, Odin processes an average of 110 sentences/second\footnote{After the initial text processing pipeline that includes syntactic parsing.} with a grammar of 211 rules on a laptop with an i7 CPU and 16GB of RAM.
 
This document is organized as follows. Section~\ref{sec:example} introduces the Odin rule language with a simple walkthrough example. Section~\ref{sec:rulelang} describes the complete rule language. The remaining sections introduce the programmatic aspects of Odin. Section~\ref{sec:mentions} describes Odin mentions, which are Scala\footnote{Odin is implemented in the Scala language. However, because Scala runs on the standard Java Virtual Machine (JVM), it plays well with other JVM languages, most notably Java.} objects that store the output of rules. Section~\ref{sec:actions} describes programmatic ways to customize the output of rules, by attaching custom Scala code to rules. An important note here is that Odin constructs these mentions automatically, so adding custom actions is completely optional and their addition should be reserved for complex phenomena that are not easily implemented with rules, e.g., coreference resolution. Lastly, Section~\ref{sec:api} puts it all together, by introducing the Odin Scala API, i.e., how to instantiate and execute a domain grammar programatically.

\section{A Walkthrough Example}
\label{sec:example}

\begin{figure}[t]
\centering
\begin{quote}
  The phosphorylation of {\bf MEK} by {\bf RAS} inhibits the ubiquitination of {\bf TGF}.
\end{quote}
\caption{A sentence containing three events in the biomedical domain: a phosphorylation, ubiquitination, and a negative regulation between the two. The text in bold marks biochemical named entities previously identified by a NER.}
\label{fig:walkthrough}
\end{figure}

Lets use the sentence in Figure~\ref{fig:walkthrough} as a simple walkthrough example for an Odin grammar in the biomedical domain. This particular sentence contains three protein named entities, previously found by a NER\footnote{Although here we focus on event extraction, Odin can also be used to write rules that extract entities. We largely ignore these type of rules here because event extraction is much more challenging and exciting.}, and we would like to build a grammar that finds three molecular events: two 
simple events that operate directly on the entities mentioned in the text, that is the phosphorylation of MEK by RAS, and the ubiquitination of TGF\footnote{It is not extremely important in this context, but, in the biomedical domain, a phosphorylation event adds a phosphate group to the corresponding protein, which alters the activity of the protein. Similarly, ubiquitination adds ubiquitin, a regulatory protein, to the corresponding substrate protein. Finally we have a more complex event
that takes these two events as arguments (phosphorylation inhibits
ubiquitination).  Detecting and linking these kinds of interactions, or``events'', deepens our understanding of cancer signaling pathways.}.

\begin{figure}[t]
\centering
\begin{lstlisting}[style=yaml-style, label={intro:rules},
    captionpos=b, caption={Rules that capture the events listed in Figure~\ref{fig:walkthrough}.}]
- name: ner
  label: Protein
  type: token
  pattern: |
    [entity="B-Protein"][entity="I-Protein"]*

- name: phospho
  label: [Phosphorylation, Event]
  pattern: |
    trigger = phosphorylation
    theme: Protein = prep_of
    cause: Protein? = prep_by

- name: ubiq
  label: [Ubiquitination, Event]
  pattern: |
    trigger = ubiquitination
    theme: Protein = prep_of
    cause: Protein? = prep_by

- name: negreg
  label: Negative_regulation
  pattern: |
    trigger = [lemma=inhibit & tag=/^V/]
    theme: Event = dobj
    cause: Event = nsubj
\end{lstlisting}
\end{figure}

Conceptually, Odin follows the same strategy introduced by FASTUS more than 20 years ago~\cite{Appelt:93}: it applies a cascade of grammars, where each grammar builds on the output produced by the previous one. This is illustrated in the grammar listed in Example~\ref{intro:rules}, which lists all the rules necessary to capture the events of interest from Figure~\ref{fig:walkthrough}. The different rules capture multiple phenomena:
\begin{description}
  \item[ner] promotes the output of the external NER, i.e., the NE labels in IOB notation\footnote{The IOB or BIO notation is a common representation, first proposed in~\cite{ramshaw1995text}, used to capture sequences of words that form named entity mentions. Please see \url{http://www.cnts.ua.ac.be/conll2003/ner/} for more examples and details.}, to Odin's mention objects, and assigns them the arbitrary label \texttt{Protein}. Note that mention labels are a domain-dependent choice, and, thus, they are the responsibility of the domain developer. The implement this rule we used a simple surface, or {\tt token}, pattern. 
  \item[phospho] matches a phosphorylation event, which is anchored around a nominal {\tt trigger}, ``phophorylation'', and has two arguments: a mandatory {\tt theme}, which is syntactically attached to the trigger verb through the preposition ``of'', and an optional (note the {\tt ?} character) {\tt cause}, attached to the trigger through the preposition ``by''. Both arguments must be {\tt Protein} mention. 
  In general, we call events that take only entity mentions as arguments {\em simple} events. 
  The resulting event mention is assigned the \texttt{Phosphorylation} and {\tt Event} labels (any number of labels $\ge 1$ can be assigned through a rule). By assigning multiple labels to a mention, a domain developer essentially implements a de facto domain taxonomy. For example, in this example, we arbitrarily decide that an {\tt IS-A} relation exists between labels from left to right. That is, the {\tt Phosphorylation} event is a type of {\tt Event}. In Section~\ref{sec:taxonomy} we discuss how to use formally-defined taxonomies in Odin.
  \item[ubiq] matches another simple event, this time around a ubiquitination. Clearly, there is a lot of redundancy between these last two rules. We will discuss later how to avoid this through rule templates. 
  \item[negreg] matches the specified trigger for a negative regulation, and then uses syntactic patterns to
    find the arguments, theme and cause, which, this time, must be event mentions. This rule will of course match only after the mentions for the simple events introduced above are constructed. We call these type of events, which take other events as arguments, {\em recursive} events.
\end{description}
Explicit priorities can be assigned to rules to control the order and extent of their execution. It is important to note that these priorities are not mandatory. If they are not specified, Odin attempts to match all rules, which imposes an implicit execution. That is, {\tt phospho} and {\tt ubiq} can only match after {\tt ner} is executed, because they require entity mentions as arguments. Similarly, {\tt negreg} matches only after the simple event mentions are constructed.

Once the domain grammar is defined, the hard work is done. 
These rules are fed into an \texttt{ExtractorEngine} Scala object that applies them on free text and returns the extracted Mentions, as summarized in Example~\ref{intro:ee}. 

\begin{figure}[t]
\centering
\begin{lstlisting}[style=scala-style, label={intro:ee},
    captionpos=b, caption={A simple Scala API example. Here we used \texttt{BioNLPProcessor}, a processor tuned for texts in the biomedical domain, for POS, NER, and syntactic analysis. We offer open-domain processors as well, such as {\tt CoreNLPProcessor}.}]
val rules = "... text containing a domain grammar ..."
// this engine applies the rules on free text and constructs output mentions
val ee = ExtractorEngine(rules)
// instantiate a Processor, for named entity recognition and syntactic analysis
val proc = new BioNLPProcessor
// annotate text, producing a document with POS, NER, and syntactic annotations
val text = "... example text ..."
val doc = proc.annotate(text)
// and, lastly, apply the domain grammar on this document
val mentions = ee.extractFrom(doc)
\end{lstlisting}
\end{figure}

Of course, this simple example does not cover all of Odin's features. In the
following sections you will learn the different features that can be
used to make more general, permissive, or restrictive rules using our
declarative language.
For advanced users, we will also demonstrate how to write custom code that can be attached to rules, also known as ``actions'', which
can be used to transform the extracted mentions in ways that are not
supported by the language, so that you can create complex systems that better
adapt to your needs.

\section{Rules}
\label{sec:rulelang}


As the previous example illustrated, the fundamental building block of an Odin grammar is a rule.  
Rules define either surface patterns, which are flat patterns over sequences of words, such as {\bf ner} in the example (formally defined in
Section~\ref{sec:tokenpattern}), or patterns over the underlying syntactic structure of a sentence described using relational dependencies, such as {\bf phospho}, {\bf ubiq}, or {\bf negreg} (defined in Section~\ref{sec:dependencypattern}).

All Odin rules are written in YAML~\cite{ben2005yaml}. However, it is not necessary to
be a YAML expert to use Odin, as we only use a small and simple YAML subset to 
write rules. A brief explanation of the required YAML features is
given in Section~\ref{sec:yaml}.

Once you are comfortable writing rules, it is time to construct a complete domain grammar. In the simplest instance, a complete grammar is a single file containing some rules (similar to Example~\ref{intro:rules}). 
While this is sufficient for simple domains, when tackling more complex
domains it may become necessary to organize rules into several files and recycle sets of prototypical rules to cover related events by altering sub-pattern variables. We describe all these situations in Section~\ref{sec:grammar}.

\subsection{A Gentle Introduction to YAML}
\label{sec:yaml}
Odin rules are written using a small YAML~\cite{ben2005yaml} subset. In particular, we
only use lists, associative arrays, and strings, which are briefly summarized below. 
For more details (although you should not need them), please read the YAML manual~\cite{ben2005yaml}.

\subsubsection{YAML Lists}
YAML supports two different ways of specifying lists. The recommended one for
Odin requires each list item to appear in a line by itself,
and it is denoted by prepending a dash and a space before the actual element. Elements of the
same list must have the same level of indentation. As an example, a list of fruits in
YAML notation is provided in Example~\ref{rules:yamllist}.

\begin{figure}[H]
\centering
\begin{lstlisting}[style=yaml-style, label={rules:yamllist},
    captionpos=b, caption={Example YAML list}]
- apple
- banana
- orange
- watermelon
\end{lstlisting}
\end{figure}

\subsubsection{YAML Associative Arrays}
\label{yaml:dict}
YAML supports two different syntaxes for associative arrays. The
recommended one for Odin is the one in which each key-value pair
appears in its own line, and all key-value pairs have the same
level of indentation. Each key must be followed by colon. 
An example of a YAML associative array is
provided in Example~\ref{rules:yamldict}.

\begin{figure}[H]
\centering
\begin{lstlisting}[style=yaml-style, label={rules:yamldict},
    captionpos=b, caption={Example YAML associative array}]
first_name: Homer
last_name: Simpson
address: 742 Evergreen Terrace
town: Springfield
\end{lstlisting}
\end{figure}

\subsubsection{YAML Strings}
Many rule components are encoded using single-line strings, as we have seen in
the previous examples. There is one exception: the rule's
\texttt{pattern} field (as described in
Sections~\ref{sec:tokenpattern}~and~\ref{sec:dependencypattern}). 
Patterns can be complex and it is a good idea to break them into
several lines. YAML supports multi-line strings using the vertical bar
character (e.g. \texttt{|}) to partition a key-value pair. When this is used, the string
begins in the next line and it is delimited by its indentation. An
example of a YAML multi-line string is shown in
Example~\ref{rules:yamlstring}.

\begin{figure}[H]
\centering
\begin{lstlisting}[style=yaml-style, label={rules:yamlstring},
    captionpos=b, caption={Example YAML associative array with one multi-line string value}]
var1: single-line string
var2: |
  this is a multi-line string
  this is still part of the same string
  because of its indentation
var3: another single-line string
\end{lstlisting}
\end{figure}

As shown, YAML strings don't have to be quoted. This is a nice
feature that allows one to write shorter and cleaner rules. 
However, there is one exception
that you should be aware of: strings that start with a YAML indicator
character must be quoted. Indicator characters have special semantics
and must be quoted if they should be interpreted as part of a string. These
are all the valid YAML indicator characters:
\begin{verbatim}
- ? : , [ ] { } # & * ! | > ' " % @ `
\end{verbatim}
As you can probably tell, these are not characters that occur
frequently in practice. Usually names and labels are composed of
alphanumeric characters and the occasional underscore, so, most of the time, you can get away without quoting strings.

\subsection{Rules}

Odin rules are represented simply as YAML associative arrays, using 
the fields shown in Table~\ref{tab:rulefields}.

\begin{table}[H]
\centering
\begin{tabular}{l|p{9cm}|l}
\textbf{Field} & \textbf{Description} & \textbf{Default} \\
\hline
\texttt{name} & The rule's name (must be unique) & \textit{must be provided} \\ 
\texttt{label} & The label or list of labels to assign to the mentions
  found by this rule & \textit{must be provided} \\ 
\texttt{priority} & The iterations in which this rule should be applied.
Note that the Odin runtime system continuously applies the given grammar on a given sentence until no new rule matches (this allows grammars that use recursive events, such as the one in Example~\ref{intro:rules} to work). Each of these distinct runs is called an ``iteration'', and they are all numbered starting from 1. Through priorities, a developer can specify in which iteration(s) the corresponding rule should run. Specifying priorities is not required, but it may have an impact on run time, by optimizing which rule should be applied when.
  A priority can be exact (denoted by a single number), a range (two
  numbers separated by a dash), an infinite range (a number followed
  by a plus \verb#+#), or a list of priorities (a comma separated list
  of numbers surrounded by square brackets. & \texttt{1+} \\ 
\texttt{action} & The custom code (or ``action'') to call for the matched mentions. As discussed in Section~\ref{sec:actions}, specifying an action is not required. The {\tt default} action does the most widely used job, i.e., keeping track of what was matched.  & \texttt{default} \\ 
\texttt{keep} & Include the output of this rule in the output
  results? & \texttt{true} \\ 
\texttt{type} & What type of rule is this: surface rule ({\tt token}) or syntax-based ({\tt dependency})? & \texttt{dependency} \\ 
\texttt{unit} & As discussed in Section~\ref{sec:tokenpattern}, each token contains multiple pieces of information, e.g., the actual word ({\tt word}), its lemma, or its part-of-speech (POS) tag ({\tt tag}). This parameter indicates which of these fields to be matched against implicitly, i.e., when the token pattern is a simple string.  
Currently, the only valid values are \texttt{word} and \texttt{tag}. & \texttt{word} \\ 
\texttt{pattern} & Either a token or a dependency pattern, as specified in \texttt{type}, that describes how to match mentions. & \textit{must be provided} \\ 
\end{tabular}
\caption{An overview of the fields of Odin rules.}
\label{tab:rulefields}
\end{table}

Clearly, the most important part of a rule, is the {\tt pattern} field. In Section~\ref{sec:tokenpattern} we describe how to implement surface, or ``token'', patterns. 
These are useful for simple sequences, or when syntax is not to be trusted.
In Section~\ref{sec:dependencypattern} we introduce the bread-and-butter of Odin: syntactic, or ``dependency'', patterns.
Note that both types of patterns use some of the same constructs: string matchers (i.e., objects that can match a string), and token constraints (i.e., objects that impose complex conditions on individual tokens to be matched). We will introduce these for token patterns, and reuse them for dependency patterns.

\subsection{Token Patterns}
\label{sec:tokenpattern}

A common task in information extraction is extracting structured
information from text. Structured information may refer to different
kinds of things, from item enumerations to complex event mentions. One
way to extract this kind of mentions from text is by the use of
surface patterns that allow us to match sequences of tokens that
usually signal the presence of the information we are interested in.

Surface patterns are available in Odin through the use of ``token''
patterns. Odin's token patterns can match continuous and discontinuous
token sequences by applying linguistic constraints on each token
(Section~\ref{sec:tokenconstraint}), imposing structure
(Section~\ref{sec:arguments}), generalized through the use of
operators (Section~\ref{sec:operators}), and drawing on context
(Section~\ref{sec:zerowidthassertions}).  In this section we will
describe each of these features that make token patterns efficient and
easy to use for the different information extraction tasks that are
encountered by practitioners.

\subsubsection{Token Constraints}
\label{sec:tokenconstraint}
Remember that, in the simplest case, a token (or word) can be matched in Odin simply by specifying a string. For example, to match the phosphorylation trigger in Example~\ref{intro:rules}, all we had to do was write {\tt phosphorylation} (quotes are optional).
But, of course, Odin can do a lot more when matching individual words. This is where token constraints become useful. 
A token constraint is a boolean expression surrounded by
square brackets that can be used to impose more complex conditions when matching a token.

Each token has multiple fields that can be matched:

\begin{table}[H]
\centering
\begin{tabular}{l|p{9cm}}
\textbf{Field} & \textbf{Description} \\
\hline
\texttt{word} & The actual token. \\
\texttt{lemma} & The lemma form of the token \\
\texttt{tag} & The part-of-speech (PoS) tag assigned to the token \\
\texttt{incoming} & Incoming relations from the dependency graph for the token \\
\texttt{outgoing} & Outgoing relations from the dependency graph for the token \\
\texttt{chunk} & The shallow constituent type (ex. \texttt{NP, VP}) immediately containing the token \\
\texttt{entity} & The NER label of the token \\
\texttt{mention} & The label of any \texttt{Mention}(s) (i.e., rule output) that contains the token. \\
\end{tabular}
\caption{An overview of the attributes that may be specified in a token constraint.}
\label{tab:tokenattribute}
\end{table}

A token field is matched by writing the field name, followed by the equals
character and a string matcher. (e.g. \verb#word=dog# matches the word
``dog'', \verb#tag=/^V/# matches any token with a part-of-speech that
starts with ``V'', \verb#entity="B-Person"# matches any token that is
the beginning of a person named entity).
Expressions can be combined using the common boolean operators: and
\verb#&#, or \verb#|#, not \verb#!#. Parentheses are also available
for grouping the boolean expressions.

Note: if the square brackets that delimit the token constraint are
left empty, i.e., {\tt []}, the expression will match any token.

\subsubsection{String Matchers}

A string matcher is an object that matches a string. Matching
strings is the most common operation in Odin, being heavily used both in token and dependency
patterns. This is because all token fields (described in Table~\ref{tab:tokenattribute})
have string values that are matched using string
matchers. 
Additionally, dependency patterns (described in
Section~\ref{sec:dependencypattern}) match incoming and outgoing dependencies
by matching the name of the dependency using the same string matchers.

Strings can be matched exactly or using regular expressions. Both
options are described next.

\subsubsection{Exact String Matchers}

An exact string matcher is denoted using a string literal,
which is a single- or double-quote delimited string. The escape
character is the backslash (e.g., \verb#\#). If the string is a valid Java
identifier, the quotes can be omitted. For example, \verb#word=dog# matches the word
``dog''. 

\subsubsection{Regex String Matchers}

A regex string matcher is denoted by a slash delimited Java
regular expression.\footnote{See
  \url{http://docs.oracle.com/javase/8/docs/api/java/util/regex/Pattern.html}}
A slash can be escaped using a backslash. This is the only escaping
done by Odin to regular expressions, everything else is handled by the
Java regular expression engine.
For example, \verb#tag=/^V/# matches any token with a part-of-speech that
starts with ``V''.

\subsubsection{Named Arguments}
\label{sec:arguments}
Token patterns support two types of named arguments: those constructed ``on-the-fly'' from an arbitrary sequence of tokens or those that point to existing mentions.  

Capturing a sequence of tokens and assigning a label to the span for later use can be performed
using the \texttt{(?<identifier> pattern)} notation, where \texttt{identifier}
is the argument name and \texttt{pattern} is the token pattern whose
result should be captured and associated with the argument name. Capturing several sequences or mentions with the same name is
supported as well as nested captures (i.e., arguments defined inside other arguments).

\begin{figure}[H]
\centering
\texttt{\textbf{Bonnie} and \textbf{Clyde} robbed the bank.}
\begin{lstlisting}[style=yaml-style,
  linerange={7}, captionpos=b, label=fig:repeatedargsonthefly, caption={An example of a token pattern with a repeated argument using a subpattern-style named argument.}]
     (?<robber> Bonnie) and (?<robber> Clyde) robbed []*? (?<location> bank)
\end{lstlisting}
\end{figure}

While powerful, these subpattern-style named arguments can quickly clutter a rule, especially when the pattern is nontrivial.  Consider the \texttt{(?$<$\textbf{robber}$>$)} pattern in Example~\ref{fig:repeatedargsonthefly}. A broad-coverage rule for detecting a \textit{robber} could be quite complex.  A better strategy might be to generalize this pattern as a rule designed to identify any \textit{person}.  Since this rule provides the context of a \textit{robbery} event, it would be sufficient to simply specify that the span of text being labelled \textit{robber} is a mention of a \textit{person}.  We can do this quite easily with Odin.

A previously matched mention can be included in a token pattern using the \verb#@# operator followed by a \texttt{StringMatcher} that
should match a mention label. This will consume all the tokens that
are part of the matched mention. If the mention should be captured in
one of the named groups then the notation is
\texttt{@identifier:StringMatcher} where the identifier is the group
name and the string matcher should match the mention label.  

\begin{figure}[H]
\centering
\texttt{\textbf{Bonnie} and \textbf{Clyde} robbed the bank.}
\begin{lstlisting}[style=yaml-style,
  linerange={7}, captionpos=b, caption={An example of a token pattern with a repeated argument using an mention-based named argument. This assumes that other rules built the {\tt Person} and {\tt Location} mentions, possibly from the output of a NER.}]
     @robber:Person and @robber:Person robbed []*? @location:Location
\end{lstlisting}
\label{fig:repeatedargs}
\end{figure}

\subsubsection{Token Pattern Operations}
\label{sec:operators}

The most fundamental token pattern operations are concatenation and
alternation. Concatenating two patterns is achieved by writing one
pattern after the other. Alternation is achieved by separating the two
patterns using the alternation operator (e.g., \verb#|#). This is
analogous to a boolean OR.

Parentheses can be used to group such expressions. As is usual, parentheses take precedence over the alternation operator.
Table~\ref{tab:operators} shows some simple examples of operator and parenthesis usage.
\begin{table}[H]
\centering
\begin{tabular}{l | l}
\textbf{pattern} & \textbf{description} \\
\hline
\verb# fat rats | mice # & matches \textit{fat rats} OR \textit{mice} \\    
\verb# fat (rats | mice) # & matches \textit{fat rats} OR \textit{fat mice} \\    
\end{tabular}
\caption{Example of parentheses usage to change operator precedence.}
\label{tab:operators} 
\end{table}


Odin also supports several types of quantifiers (see
Table~\ref{tab:quantifiers} for details).  The \verb#?#, \verb#*# and
\verb#+# postfix quantifiers are used to match a pattern zero or one
times, zero or more times, and one or more times respectively. These
are greedy quantifiers, and can be turned lazy by appending a question
mark (e.g., \verb#??#, \verb#*?#, \verb#+?#).
Figure~\ref{fig:greedylazy} illustrates the difference between greedy
and lazy quantifiers.

\begin{figure}[H]
\centering
\texttt{a b c d e f c}
\begin{table}[H]
\centering
\begin{tabular}{l | l}
\textbf{pattern} & \textbf{match} \\
\hline
\verb#[]+ c# & a b c d e f c \\
\verb#[]+? c# & a b c \\
\end{tabular}    
\end{table}
\caption{Comparison of greedy (default behavior) and lazy ({\tt ?}) quantifiers.}
\label{fig:greedylazy}
\end{figure}

Ranged repetitions can be specified by appending \verb#{n,m}# to a
pattern, which means that the pattern should repeat at least $n$ times
and at most $m$. If $n$ is omitted (e.g., \verb#{,m}#) then the
pattern must repeat zero to $m$ times. If $m$ is omitted (e.g.,
\verb#{n,}#) then the pattern must repeat at least $n$ times. Ranges
are greedy, and can be turned lazy by appending a question mark (e.g.,
\verb#{n,m}?#, \verb#{,m}?#, \verb#{n,}?#)
For an exact number of repetitions the \verb#{n}# suffix is provided.
Since this is an exact repetition there are no greedy/lazy variations.

Table~\ref{tab:quantifiers} summarizes this set of quantifiers.

\begin{table}[H]
\centering
    \begin{tabular}{l| p{8cm} | l}
    \textbf{Symbol} & \textbf{Description} & \textbf{Lazy form} \\
    \hline
    \texttt{?} & The quantified pattern is optional. & \texttt{??} \\
    \texttt{*} & Repeat the quantified pattern \textit{zero} or more times. & \texttt{*?} \\
    \texttt{+} & Repeat the quantified pattern \textit{one} or more times. & \texttt{+?} \\
    \verb#{n}# & Exact repetition.  Repeat the quantified pattern $n$ times. & \\
    \verb#{n,m}# & Ranged repetition.  Repeat the quantified pattern between $n$ and $m$ times, where $n < m$. & \verb#{n,m}?# \\
    \verb#{,m}# & Open start ranged repetition.  Repeat the quantified pattern between 0 and $m$ times, where $m > 0$. & \verb#{,m}?# \\
    \verb#{n,}# & Open end ranged repetition.  Repeat the quantified pattern at least $n$ times, where $n > 0$. & \verb#{n,}?# \\
    \end{tabular}    
    \caption{An overview of the quantifiers supported by Odin's token patterns.
}
    \label{tab:quantifiers}
\end{table}

Quantifiers apply either to a single token constraint or to a group of
token constraints. Groups are specified by using parentheses. An
example of a token pattern that uses quantifiers is shown on
Example~\ref{tokpat:quantifiers}. This example also shows that one can
use mention captures in the quantified groups (the {\tt Number} argument), and that the captured
mentions can share the same name. This is useful for the extraction of
enumerations of unknown length.

\begin{figure}[H]
\centering

\texttt{The numbers \textbf{4}, \textbf{8}, \textbf{15}, \textbf{16}, \textbf{23} and \textbf{42} frequently recurred in Lost.}

\begin{lstlisting}[style=yaml-style, label={tokpat:quantifiers},
    captionpos=b, caption={Example showcasing quantifiers and mention captures.}]
# First, find numbers by inspecting the POS tag.
# Note that this is not the only way to check for a number,
# there are other options, such as [word=/\d+/]
- name: numbers
  label: Number
  priority: 1
  type: token
  pattern: |
    [tag=CD]

# Second, match comma separated lists of numbers optionally followed
# by the word 'and' and a final number.
- name: list
  label: ListOfNumbers
  priority: 2
  type: token
  pattern: |
    @num:Number ("," @num:Number)+ (and @num:Number)?
\end{lstlisting}
\end{figure}

\subsubsection{Zero-width Assertions}
\label{sec:zerowidthassertions}

Zero-width assertions allow one to verify whether or not a pattern is
present without including it in the matched result.  
Odin supports the following zero-width assertions:

\begin{table}[H]
    \begin{tabular}{l| l | l}
    \textbf{Symbol} & \textbf{Description} & \textbf{Limitation} \\
    \hline
    \texttt{\^} & beginning of sentence & \\
    \texttt{\$} & end of sentence & \\
    \texttt{(?=\dots)} & postive lookahead & \\
    \texttt{(?!\dots)} & negative lookahead & \\ [1ex]
    \hline 
    & & \\
    \texttt{(?<=\dots)} & positive lookbehind & 
    \multirow{2}{8cm}{The length of the lookbehind assertion matches
      must be known at compile time. This restricts the patterns
      supported by lookbehinds to token constraints and the exact
      range quantifier (e.g., \texttt{\string{n\string}}). Parentheses
      are supported.
    } \\ [5ex]
    \texttt{(?<!\dots)} & negative lookbehind & \\
    \end{tabular}    
    \caption{An overview of the zero-width assertions supported by Odin.  
    These patterns do not consume tokens, but are useful to match patterns preceding/following the expression of interest.}
\end{table}

\subsubsection{Output}
The output of any Odin rule is called a ``mention'', and they are actual instances of a {\tt Mention} Scala class, or one of its subclasses (see Section~\ref{sec:mentions}). 

The inclusion of named captures in a token pattern affects the type of \texttt{Mention}
that is produced.  In general, the result of applying a token pattern
successfully is usually a \texttt{TextBoundMention} (see Section~\ref{sec:mentions}). 
However, if the
token pattern includes named captures, then the result is a
{\tt RelationMention}, which is essentially a collection of named captures, or ``arguments'' (but without a predicate, or ``trigger'', which is typical of event mentions!).
In other words, relation mentions are not dependent on a particular predicate.  
If one of the named captures has the name ``trigger'' (case insensitive), then Odin assumes that this pattern defines an event, and the result is an event mention (an instance of the {\tt EventMention} class). Examples~\ref{tokpat:eventout} and~\ref{tokpat:relout} show two simple patterns that produce an event mention and a relation mention, respectively.

\begin{figure}[H]
\centering

\texttt{\textbf{Oscar} lives in a \textbf{trash can}.}

\begin{lstlisting}[style=yaml-style, label={tokpat:eventout},
    captionpos=b, caption={An example of a token pattern rule that produces an event mention through the specification of a trigger.}]
- name: event_mention_out
  label: LivesIn
  priority: 2
  type: token
  pattern: |
    (?<resident>Oscar) 
    (?<trigger>[lemma=live]) 
    in [tag=DT]? (?<location>[tag=/^N/]+)
\end{lstlisting}
\end{figure}

\begin{figure}[H]
\centering

\texttt{\textbf{Dr.} \textbf{Frankenstein} spends a lot of time in the graveyard.}

\begin{lstlisting}[style=yaml-style, label={tokpat:relout},
    captionpos=b, caption={An example of a token pattern rule that produces a relation mention. This rule has named arguments, but does not specify a trigger.  For brevity, we assume that \texttt{Person} mentions have already been identified.}]
- name: relation_mention_out
  label: PersonWithTitle
  priority: 2
  type: token
  pattern: |
    (?<title>[word=/(?i)^mr?s|dr|prof/]) @person:Person
\end{lstlisting}    
\end{figure}

\subsection{Dependency Patterns}
\label{sec:dependencypattern}

While token patterns are quite powerful, they are, of course, not too robust to syntactic variation.  Writing patterns over syntactic structure produces generalizations with broader coverage that do not sacrifice precision.  Consider the sentences in Figure~\ref{fig:syntacticvariation}:

\begin{figure}[H]
\begin{itemize}
\item[] \texttt{\textbf{Noam} \textbf{danced} at midnight with the \textbf{leprechaun}.}
\item[] \texttt{\textbf{Noam}, in full view of the three-legged robot, \textbf{danced} at dawn with the \textbf{leprechaun}.}
\item[] \texttt{\textbf{Noam} \textbf{danced} under the moonlight at midnight with the \textbf{leprechaun}.}
\item[] \texttt{His friends watched in awe while \textbf{Noam} \textbf{danced} the forbidden jig with the \textbf{leprechaun} at midnight.}
\end{itemize}
\caption{These sentences show some of the infinite syntactic variation describing a dance between two entities.}
\label{fig:syntacticvariation}    
\end{figure}

\begin{figure}[H]
\centering
\includegraphics[width=\linewidth]{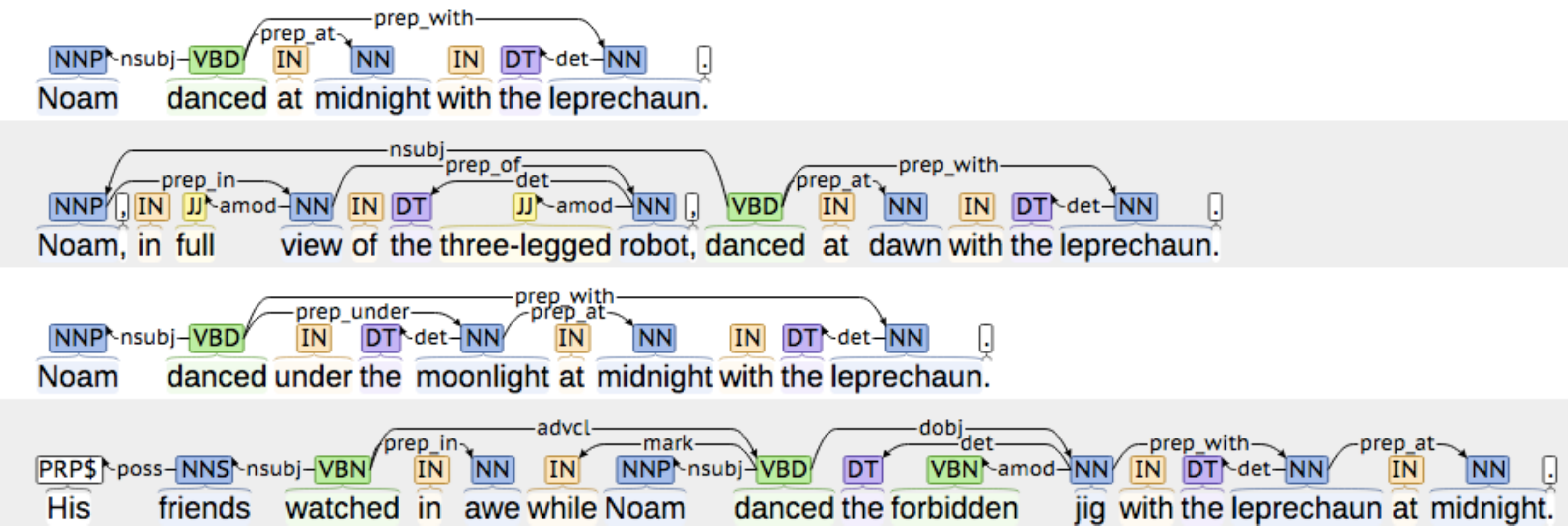}
\caption{The relational-dependency parse for the sentences in Figure~\ref{fig:syntacticvariation}.}
\label{fig:noamsyntax}
\end{figure}

While it requires several token-pattern rules to precisely capture the syntactic variation shown in Figure~\ref{fig:syntacticvariation}, all of these variants can be covered with a single rule using a dependency pattern (see Example~\ref{deppat:syntacticvariation}).

\begin{lstlisting}[style=yaml-style, label={deppat:syntacticvariation},
    captionpos=b, caption={A dependency rule that expects two arguments: (1) a nominal subject and (2) the head word complements of a ``with'' prepositional phrase off of the lemmatized trigger, \textit{dance}; (2) may be preceded by an optional hop through a direct object (\texttt{dobj}) relation.  Note the optional hop through a direct object (\texttt{dobj}). Parsers often struggle with prepositional attachment, so we have added an optional \texttt{dobj} in this rule to be robust to such errors.}]
- name: dancers_1
  label: Dance
  priority: 2
  pattern: |
    trigger = [lemma=dance]
    dancer:Entity = nsubj
    partner:Entity = dobj? prep_with
\end{lstlisting} 

\begin{figure}[H]
\centering
\includegraphics[width=\linewidth]{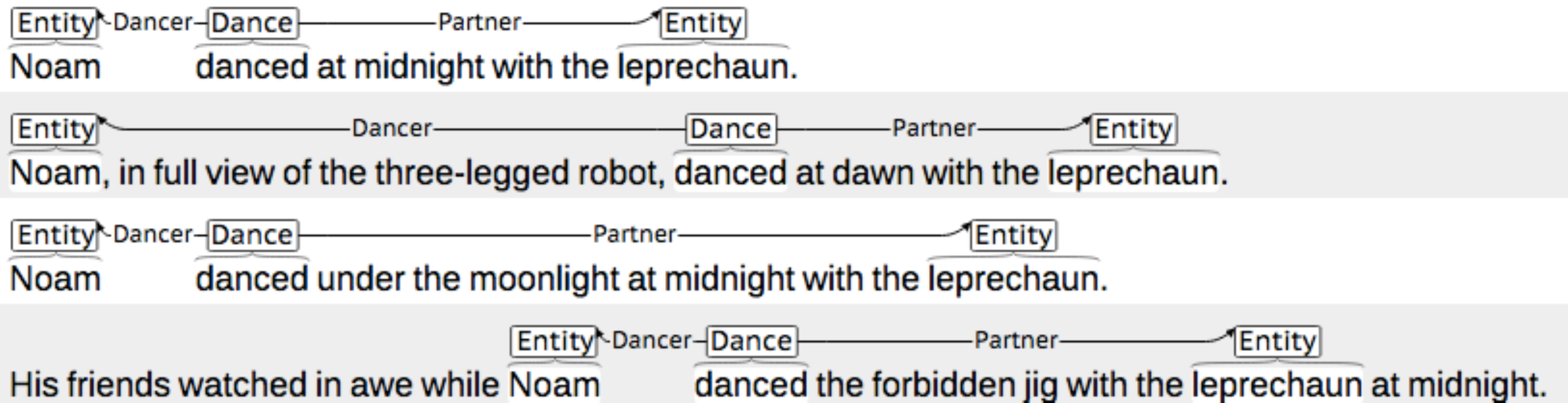}
\caption{The structured output of the rule in Example~\ref{deppat:syntacticvariation}.}
\end{figure}

Formally, a dependency pattern describes a traversal over a syntactic
dependency graph.  
Again, we currently use Stanford dependencies~\cite{Marneffe:08} in Odin, 
but Odin is independent of the representation used.
Odin's dependency patterns are composed of several fields.
To boot, dependency patterns defining event rules require a ``trigger'' that
must be set to a token pattern (see previous section).  
This token pattern describes a valid predicate for the event of interest.  
The rest of the
fields are event arguments defined through a syntactic path from the trigger
to some mention (entity or event) that was previously matched by another rule.  
The path is composed of \textit{hops} and
optional \textit{filters}.  The hops are edges in the syntactic
dependency graph; the filters are token constraints on the
nodes (tokens) in the graph.


Hops can be \textit{incoming} or \textit{outgoing}.  An \textit{outgoing} hop follows the
direction of the edge from \textsc{head}
$\rightarrow$\textsc{dependent}; an \textit{incoming} hop goes against the
direction of the edge, leading from \textsc{dependent}
$\rightarrow$\textsc{head}. For example, the dependency ``jumped'' $\rightarrow$ ``Fonzie'' is outgoing (``jumped'' is the head), but it is considered incoming when traversed in the other direction: ``Fonzie'' $\leftarrow$ ``jumped''.

\begin{figure}[H]
\centering
\label{fig:dep-graph}
%
\includegraphics[scale=0.6]{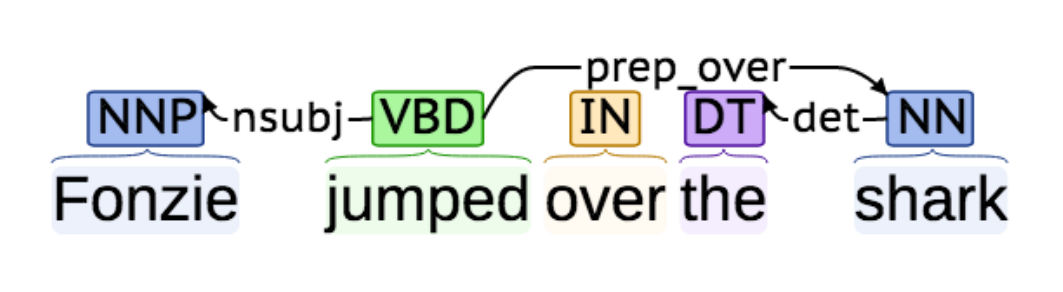}
\caption{A simple sentence with its corresponding dependency parse tree.}
%
%
\end{figure}

\begin{figure}[H]
\centering
\begin{lstlisting}[style=yaml-style, linerange={12-15}, captionpos=b, label=fonzie,
caption={A simple, two-argument dependency pattern composed solely of outgoing hops, which matches the ``jumping'' event above. We are assuming that a different rule created a {\tt Noun} mention for every {\tt NN*}.}]
  pattern: |
    trigger = [lemma=jump]
    entity:Noun = nsubj
    obstacle:Noun = prep_over
\end{lstlisting}
\end{figure}

An outgoing dependency is matched using the \texttt{>} operator
followed by a string matcher, which operates on the label of the corresponding dependency, e.g., \texttt{>nsubj}. 
Because most patterns use outgoing hops, 
(i.e. \textsc{head} $\rightarrow$\textsc{dependent}), the \texttt{>}
operator is implicit and can therefore be omitted.  An incoming
relation (i.e. \textsc{dependent} $\rightarrow$\textsc{head}) is
matched using a required \texttt{<} operator followed by a
string matcher. \texttt{>>} is a wildcard operator that can be
used to match any outgoing dependency.  \texttt{<<} is a wildcard
operator that can be used to match any incoming dependency.


Importantly, restrictions may be imposed on the nodes (i.e., tokens) visited when
following dependencies, using the usual token constraints.
Example~\ref{ex:steal} illustrates such constraints on both the {\tt robber} (using the POS tag) and the {\tt property} (using the actual word). 

\begin{figure}[H]
\centering
\label{rule:steal}
\begin{itemize}
\item[] \texttt{\textbf{Gonzo} \textbf{stole} her \textbf{chicken}.}
\item[] \texttt{\textbf{Gonzo} \textbf{stole} her \sout{heart}.}
\end{itemize}
\begin{lstlisting}[style=yaml-style, captionpos=b, label=ex:steal,
  caption={While these two sentences are syntactically identical, only one pertains to theft of tangible goods. We are assuming that a different rule created a {\tt Noun} mention for every {\tt NN*}.}]
- name: np
  label: Noun
  priority: 1
  type: token
  unit: tag
  pattern: |
    /^N/

- name: steal-1
  label: Steal
  priority: 2
  pattern: |
    trigger = [lemma=steal]
    robber:Noun = nsubj [tag=NNP] # We are only interested in Proper Nouns
    property:Noun = dobj [!word=heart] # Let's keep the romance out of it.
\end{lstlisting}
\end{figure}

Just as in token patterns, dependency patterns can
include parentheses and the alternation operator \verb#|#.
For example, the pattern \texttt{nsubj|agent} matches an outgoing dependency whose label 
is either \texttt{nsubj} or \texttt{agent}.

\subsubsection{Named Arguments for Dependency Patterns}

Clearly, naming event arguments is important (e.g., one may want to keep track who is the agent and who is the patient in a robbery event). 
We probably already observed that Odin has a simple syntax for this:
a path to an argument begins with \texttt{name:label = path}, where
\texttt{label} is the the label of an existing \texttt{Mention}.  The
path must lead to a token that is a part of a \texttt{Mention} with
the specified label.
Argument names are required and \textit{unique}, i.e., you can't have two different patterns with the same name. But the same pattern may match multiple mentions!
If, for example, an argument with the name ``theme'' matched three different
entities, then three event mentions will be generated, each
with one entity as the \textit{theme}. If the given path to the \textit{theme} fails to match any
\texttt{Mention}, then no event mentions will be created. 

At times one may need to make an argument optional or allow for more
than one argument with the same name in a single
event mention.  This can be achieved through the use of
argument quantifiers.  Arguments can be made optional with the
\texttt{?} operator. The \verb#+# operator is used to indicate that a
single event mention with all matches should be created. The
\verb#*# is similar to \verb#+# but also makes the argument optional.
If the exact number of arguments with the same name is known, it can
be specified using the exact repetition quantifier \texttt{\{$k$\}}.
In cases of exact repetitions, the cartesian product will be applied
to the \texttt{Mention}s matching the given path.  If $k$
\texttt{Mention}s are asked for in a path $p$ and $n$ are found to
match $p$, then $j$ event mentions will be produced, where $j$
is the binomial coefficient shown in Equation~\ref{eq:nchoosek}.
A few rules using these argument quantifiers are shown in Examples~\ref{ex:no-quant},~\ref{ex:plus-quant}, \&~\ref{ex:k-quant}.

\begin{equation}
\binom{n}{k} = \frac{n!}{k! (n - k)!}
\end{equation}
\label{eq:nchoosek}

\begin{figure}[H]
\centering
\texttt{\textbf{Cities} like \textbf{London}, \textbf{Paris}, \textbf{Tokyo}, and \textbf{Beijing}.}
\caption{The sentence used by the rules shown in Examples~\ref{ex:no-quant}, \ref{ex:plus-quant}, \& \ref{ex:k-quant}.  We are assuming that a different rule created a {\tt Location} mention for every location NE.}
\label{fig:arg-quants}
\end{figure}

\begin{lstlisting}[style=yaml-style, captionpos=b, linerange={1-7}, label=ex:no-quant,
  caption={An example of a complete dependency pattern rule without an argument quantifier.}]
- name: cities-1
  label: Cities
  priority: 2
  pattern: |
    trigger = [lemma=city]
    # produces 4 EventMentions each with 1 city
    city:Location = prep_like
\end{lstlisting}
\begin{table}[H]
    \centering
    \begin{tabular}{l | l}
    \textbf{Mention} & \textbf{Cities} \\
    \hline
    1 & London \\
    2 & Paris \\
    3 & Tokyo \\
    4 & Beijing \\
    \end{tabular}
\caption{The four mentions produced by the dependency pattern shown in Example~\ref{ex:no-quant}.}
\label{tab:basic-citiesout}
\end{table}

\begin{lstlisting}[style=yaml-style, captionpos=b, linerange={13-15}, label=ex:plus-quant,
  caption={An example of a dependency pattern with a $+$ argument quantifier. Its output is shown in Table~\ref{tab:plus-citiesout}.}]
    trigger = [lemma=city]
    # produces 1 EventMention with 4 cities
    city:Location+ = prep_like
\end{lstlisting}
\begin{table}[H]
    \centering
    \begin{tabular}{l | l}
    \textbf{Mention} & \textbf{Cities} \\
    \hline
    1 & London, Paris, Tokyo, Beijing \\
    \end{tabular}
\caption{The single mention produced by the rule shown in Example~\ref{ex:plus-quant}.}
\label{tab:plus-citiesout}
\end{table}

\begin{lstlisting}[style=yaml-style, captionpos=b, linerange={21-25}, label=ex:k-quant,
  caption={An example of a dependency pattern with a {\tt \{k\}} quantifiers on event arguments. The scattering effect of the $\{k\}$ quantifier is shown in Table~\ref{tab:k-citiesout}.}]
    trigger = [lemma=city]
    # produces 6 EventMentions each with 2 cities
    city:Location{2} = prep_like
\end{lstlisting}
\begin{table}[H]
    \centering
    \begin{tabular}{l | l}
    \textbf{Mention} & \textbf{Cities} \\
    \hline
    1 & London, Paris \\
    2 & London, Tokyo \\
    3 & London, Beijing \\
    4 & Paris, Tokyo \\
    5 & Paris, Beijing \\
    6 & Tokyo, Beijing
    \end{tabular}
    \caption{The six mentions produced by the rule shown in Example~\ref{ex:k-quant}.}
    \label{tab:k-citiesout}
\end{table}

%

\subsubsection{Quantifiers for Dependency Patterns}

In addition of the above quantifiers on event arguments, Odin supports
quantifiers inside the actual dependency patterns. They are shown in
Table~\ref{tab:dep-quantifiers}.

The \verb#?#, \verb#*# and \verb#+# postfix quantifiers are used to
match a pattern zero or one times, zero or more times, and one or more
times respectively. There is no notion of greedy/lazy
dependency patterns.

Ranged repetitions can be specified by appending \verb#{m,n}# to a
pattern, and means that the pattern should repeat at least $m$ times
and at most $n$. If $m$ is omitted (e.g., \verb#{,n}#) then the
pattern must repeat zero to $n$ times. If $n$ is omitted (e.g.,
\verb#{m,}#) then the pattern must repeat at least $m$ times. There is
no notion of greedy/lazy dependency patterns.
For an exact number of repetitions the \verb#{n}# suffix is provided.

For example, the pattern {\tt /prep/+} matches a sequence of 1 or more
outgoing dependencies whose labels contain {\tt prep}. The pattern
{\tt dobj* /prep/\{,3\}} matches 0 or more {\tt dobj} dependencies,
followed by up to 3 outgoing dependencies that contain {\tt prep}.

\begin{table}[H]
\centering
    \begin{tabular}{l| p{8cm}}
    \textbf{Symbol} & \textbf{Description} \\
    \hline
    \texttt{?} & The quantified pattern is optional. \\
    \texttt{*} & Repeat the quantified pattern \textit{zero} or more times. \\
    \texttt{+} & Repeat the quantified pattern \textit{one} or more times. \\
    \verb#{n}# & Exact repetition.  Repeat the quantified pattern $n$ times. \\
    \verb#{n,m}# & Ranged repetition.  Repeat the quantified pattern between $n$ and $m$ times, where $n < m$. \\
    \verb#{,m}# & Open start ranged repetition.  Repeat the quantified pattern between 0 and $m$ times, where $m > 0$. \\
    \verb#{n,}# & Open end ranged repetition.  Repeat the quantified pattern at least $n$ times, where $n > 0$. \\
    \end{tabular}
    \caption{An overview of the quantifiers supported by Odin's dependency patterns.}
    \label{tab:dep-quantifiers}
\end{table}

\subsubsection{Zero-width Assertions}
For dependency patterns, there no lookbehind or lookahead assertions, only lookaround assertions.
The lookaround syntax is \texttt{(?= pattern)} for positive assertions and \texttt{(?! pattern)} for negative assertions. Example~\ref{ex:lookaround} shows an example of a positive lookaround in action.

\begin{figure}[H]
\centering
\label{rule:lookaround}
\begin{itemize}
\item[] \texttt{\textbf{Dennis} \textbf{crashed} his mom's \textbf{car}.}
\item[] \texttt{\textbf{Dennis} \textbf{crashed} his \sout{car}.}
\end{itemize}
\begin{lstlisting}[style=yaml-style, captionpos=b, label=ex:lookaround,
  caption={Sometimes ownership matters.  Perhaps we want to know whether or not Dennis should be grounded for crashing a car.  Did Dennis crash his mother's car? A positive lookaround is needed for this.}]
- name: np
  label: Noun
  priority: 1
  type: token
  unit: tag
  pattern: |
    /^N/

- name: accident-1
  label: Accident
  priority: 2
  pattern: |
    trigger = [lemma=crash]
    agent:Noun = nsubj [tag=NNP] # We are only interested in Proper Nouns
    # Only match if this is mom's car
    vehicle:Noun = dobj (?= poss [lemma=mom]) [lemma=car]
\end{lstlisting}
\end{figure}

\subsubsection{Output}
The result of applying a dependency pattern successfully is
usually an event mention.  If a trigger is not specified, a
relation mention is produced (see Figures~\ref{deppat:eventout}
\&~\ref{deppat:relout} for details).


\begin{figure}[H]
\centering

\texttt{\textbf{Oscar} lives in a \textbf{trash can}.}

\begin{lstlisting}[style=yaml-style, label={deppat:eventout},
    captionpos=b, caption={An example of a dependency pattern rule that produces an event mention through the specification of a trigger.}]
- name: dep_event_mention_out
  label: LivesIn
  priority: 2
  pattern: |
    trigger = [lemma=live]
    resident:Person = nsubj
    location:Location prep_in
\end{lstlisting}
\end{figure}

\begin{figure}[H]
\centering

\texttt{\textbf{Dr.} \textbf{Frankenstein} spends a lot of time in the graveyard.}
\begin{lstlisting}[style=yaml-style, label={deppat:relout},
    captionpos=b, caption={An example of a dependency pattern rule that produces a relation mention. This rule has named arguments, but does not specify a trigger.  When the trigger field is omitted in a dependency pattern, the first field given should specify a named argument using the mention retrieval syntax (\texttt{argname:MentionLabel}). All subsequent dependency patterns used by the other arguments are anchored on this first argument.
}]
- name: sometitle-1
  label: Title
  priority: 1
  type: token
  pattern: |
    [word=/(?i)^mr?s|dr|prof/]
    
- name: dep_relation_mention_out
  label: PersonWithTitle
  priority: 2
  pattern: |
    person:Person
    title:Title = nn
\end{lstlisting}    
\end{figure}

\subsection{Building a Grammar}
\label{sec:grammar}

By now, we hope you are somewhat confident that you can write Odin rules.
Of course, the next step is to put them together into a complete grammar.
This can be very simple: minimally, all you have to do is to store them all into a single file which is then loaded into an Odin engine (see Section~\ref{sec:api}). If you care a lot about efficiency, you can tune your grammar by assigning priorities to rules. For example, rules that match entities should be executed before (i.e., have a lower priority) than rules that match events where these entities serve as arguments. (But again, this is not needed: Odin takes care of pipelining rules internally.)  

But some domain grammars are more complicated than a simple sequence of rules. You may have event labels that are so complex that you would prefer to store them in a taxonomy. Some event types have almost exactly the same syntactic representations as others, so you would like to reuse some rules. Odin supports all these issues. We describe them next.

\subsubsection{Master File}

The master file is a grammar's entry point, or the file is passed to the Odin runtime engine.
As mentioned, for simple grammars, this file can be simply a collection of rules.
For more complicated scenarios, this file must contain 
a required \texttt{rules} section, and two optional
sections: \texttt{taxonomy} and \texttt{vars}.
Let us describe these sections next.

\subsubsection{Taxonomy}
\label{sec:taxonomy}
The taxonomy is a forest (meaning a collection of trees) of labels that, if specified, is used by Odin as the hierarchy for mention labels. An example taxonomy is shown in
Example~\ref{rules:taxonomy}.

\begin{figure}[H]
\centering
\begin{lstlisting}[style=yaml-style, label={rules:taxonomy},
    captionpos=b, caption={Example taxonomy}]
# a tree hierarchy can be used to define the taxonomy
- organism:
  - prokaryotic:
    - archaebacteria
    - eubacteria
  - eukaryotic:
    - unicellular:
      - protista
    - multicellular:
      - autotrophic:
        - plantae
      - heterotrophic:
        - fungi
        - animalia
# we want to include robots in our taxonomy
# but they are not organisms, what can we do?
# fortunately, multiple trees are supported
- robot
\end{lstlisting}
\end{figure}

If a taxonomy is provided, then all the labels used by the rules must
be declared in the taxonomy. This is obviously useful for catching typos. 
More importantly, the taxonomy hierarchy is used to robustly match mentions in subsequent
rules.  For example, if a rule creates an entity mention with the label {\tt animalia} from the taxonomy in Example~\ref{rules:taxonomy}, this mention will be matched as argument in a subsequent rule, which requires that argument to be of label {\tt multicellular}. This is because {\tt animalia} is a hyponym of {\tt multicellular}, i.e., it is directly derived from it. 

If the value of \texttt{taxonomy} is a single string, then it will be
interpreted as a file name and the taxonomy will be read from that
file.  It should be noted that the taxonomy may only be specified in the master file, whether included directly or provided through an import (see Example~\ref{fig:taxonomy-import}).

\begin{figure}[H]
\centering
\begin{lstlisting}[style=yaml-style, captionpos=b, label={fig:taxonomy-import}, caption={An example of a taxonomy import.}]
# the taxonomy file should contain only the contents of the taxonomy (without the taxonomy: section name)
taxonomy: path/to/taxonomy.yml
  
\end{lstlisting}
\end{figure}

\subsubsection{Variables and Templates}

It is very common that similar events share the same syntactic structure. For example, in the biomedical domain, all the biochemical reactions (there are between 10 and 20 of these) share the same structure. For example, ``A phosphorylates B'' is similar to ``A ubiquitinates B'', with the exception of the predicate: ``phosphorylates'' vs. ``ubiquitinates''. In such situations, we would like to reuse these syntactic structures between events (so we do not write the same rules 10--20 times). Odin supports these through the use of variables and rule templates, where rule templates are simply rules that use variables. For example, one could write a single rule template for the above example, where the trigger constraints are encoded using a variable.

In general, variables can be declared as a YAML mapping, and they can be
substituted in rules using the \verb#${variableName}# notation (see Examples~\ref{rules:masterfileex} \&~\ref{rules:importedfile}).
Furthermore, wherever a rule can be specified, you can also import a
file, through the \texttt{import}
command, and its optional \texttt{vars} parameter. 
This gives one a further opportunity to instantiate variables. 
Example~\ref{rules:masterfileex} shows the \texttt{import} command in action.

\begin{figure}[H]
\centering
\begin{lstlisting}[style=yaml-style, label={rules:masterfileex},
    captionpos=b, caption={An example of a master file that uses import statements and demonstrates variable precedence. Note that variables can be instantiated in three different places: (a) in the template file itself, (b) when the {\tt import} command is used, or (c) at the top of the master file. The precedence is: (b) $>$ (c) $>$ (a). For this particular example, it means that the value chosen for the {\tt myTrigger} variable is ``eat'' for the first import (\textsc{line} 9) and ``sell'' for the second import (\textsc{line} 13).}]
# global variables
vars:
  myTrigger: "eat"

rules:
  # import rules from file
  # if variables are used in the imported file, 
  # they will be retrieved from the global vars
  - import: path/to/template.yml

  # import rules from file
  # myTrigger is overridden for this import
  - import: path/to/template.yml
    vars:
      myTrigger: "sell"

  # rules and imports can live together in harmony :)
  - name: somedude
    label: Person
    type: token
    pattern: |
      [entity='B-Person'] [entity='I-Person']*
\end{lstlisting}
\end{figure}

\begin{figure}[H]
\centering
\begin{lstlisting}[style=yaml-style, label={rules:importedfile},
    captionpos=b, caption={The \texttt{template.yml} file imported in Example~\ref{rules:masterfileex}.}]
vars:
  # these variables are superseded by those in the master file
  myTrigger: "buy"
  myMentionLabel: "Food"
  
rules:
 - name: example_rule
   label: Event
   type: token
   priority: 1
   pattern: |
     @person:Person # match a person
     (?<trigger> [lemma=${myTrigger}]) # trigger comes from provided variable
     [tag="DT"]? @object:${myMentionLabel} # retrieve mention with given label

\end{lstlisting}
\end{figure}
\section{Mentions, or the Output of Rules}
\label{sec:mentions}

As hinted before in this document, each rule produces a {\tt Mention} object when it successfully matches some text. These objects are nothing magical: we simply use them to store everything that the rule contains, and the corresponding text matched. 
Table~\ref{tab:mentionfields} summarizes the fields of the mention object.

\begin{table}[H]
\centering
\begin{tabular}{l|p{9cm}}
\textbf{Field} & \textbf{Description} \\
\hline
\texttt{labels} &  The sequence of labels to associate with a mention. \\
\texttt{tokenInterval} &   The open interval token span from the first word to the final word $+ 1$. \\
\texttt{startOffset} & The character index in the original text at which the mention begins.  \\
\texttt{endOffset} & The character index in the original text at which the mention ends $+ 1$.  \\
\texttt{sentence} & The sentence index of this mention. \\
\texttt{document} & The document (composed of annotated sentences) from which this mention originates.  \\
\texttt{arguments} & A map from argument name (a \texttt{String}) to a sequence of mentions. \\
\texttt{foundBy} & The name of the rule that ``found'' this mention. \\
\end{tabular}
\caption{An overview of the most important fields of the {\tt Mention} class.}
\label{tab:mentionfields}
\end{table}

Sometimes, actual code is best at explaining things. We highly encourage the reader to take a look at the code implementing {\tt Mention} and its subclasses\footnote{\url{https://github.com/clulab/processors/blob/master/src/main/scala/edu/arizona/sista/odin/Mention.scala}}. Note that some the information stored in mentions, e.g., the token interval of the mention, refer to data structures produced by our preprocessing code, such as {\tt Sentence} and {\tt Document}. Again, reading through the code is the best way to learn about these\footnote{\url{https://github.com/clulab/processors/blob/master/src/main/scala/edu/arizona/sista/processors/Document.scala}}.

\subsection{TextBoundMention}

The {\tt Mention} class is subclassed by several other classes. The simplest is \texttt{TextBoundMention}.
A \texttt{TextBoundMention} is created when the output of a rule is a flat structure, i.e., a contiguous sequence of tokens in a sentence. More formally, a \texttt{TextBoundMention} will have a {\tt tokenInterval}, but its {\tt arguments} map will be empty. These mentions are usually used to represent entities or event triggers.

\subsection{RelationMention}

A {\tt RelationMention} encodes $n$-ary relations between its arguments.
All the arguments are named (based on the argument names specified in the matched rule), and are stored in the {\tt arguments} map. Importantly, several arguments may have the same name! This is extremely useful when one needs to capture in  a rule enumerations of several valid arguments with the same role (for example, a {\tt food} argument may capture multiple foods consumed at a dinner).

%
%
%

\subsection{EventMention}
 An {\tt EventMention} is similar to a
{\tt RelationMention}, with just one additional feature: it has a
\texttt{TextBoundMention} that represents the trigger of the event.
In other words, the {\tt arguments} map contains an additional argument, labeled {\tt trigger}, which stores the event's predicate. 
Note: an event must have exactly one trigger.

\section{Advanced: Customizing Rule Output with Actions}
\label{sec:actions}

Note: you are welcome to skip this section. We expect only a small numbers of users, who need deep customization of Odin, to find this section necessary.\\

As described in the previous section, Odin rules produce mentions, which store all the relevant information generated during the match. This is sufficient for most common usages of Odin, but sometimes this information requires some changes. For example, one could use coreference resolution to replace event arguments that are pronouns with their nominal antecedents, as indicated by the coreference resolution component. This is not easily done though rules, and this is when actions become necessary.

Actions are Scala methods (implemented by the domain developer!) that can be applied by Odin's runtime engine to the resulting \texttt{Mention}s after matching the rule. 
An \texttt{Action} has the type signature shown in
Example~\ref{scala:action-sig}.

\begin{figure}[H]
\centering
\begin{lstlisting}[style=scala-style, label={scala:action-sig},
    captionpos=b, caption={Signature of action methods.}]
def action(mentions: Seq[Mention], state: State): Seq[Mention]
\end{lstlisting}
\end{figure}

A rule will first try to apply its pattern to a sentence.  Any matches will be sent to the corresponding action as a \texttt{Mention} sequence.  If an action is not explicitly named, the default identity action will be used, which returns its input mentions unmodified (i.e. the input's identity).Actions receive as input parameters this \texttt{Mention} sequence and also the runtime engine's \texttt{State}.

 The {\tt State} object (second parameter) provides read-only access to {\em all} the mentions previously created by Odin in the current document. This information may be useful to implement global decisions, e.g., coreference resolution across the entire document.

Actions must return a \texttt{Mention} sequence that will be added to the \texttt{State} at the
end of the current iteration by the runtime engine. For example, the simplest possible action would return the {\tt mentions} it received as the first input parameter. Example~\ref{scala:example-action} shows an only slightly more complicated action that removes any \texttt{Mention} containing the text ``Fox''.

\begin{figure}[H]
\centering
\begin{lstlisting}[style=scala-style, label={scala:example-action},
    captionpos=b, caption={An example of an \texttt{action} that removes any \texttt{Mention} containing the text ``Fox''.}]
def action(mentions: Seq[Mention], state: State): Seq[Mention] = {
    mentions.filter(_.text contains "Fox")
}
\end{lstlisting}
\end{figure}

Note that, in addition to attaching actions to individual rules,  
actions can also be called globally at the end of each iteration by the runtime engine. 
This means that the extractor engine (see Example~\ref{scala:engine-variants}) must receive this global action as a parameter during its construction. 

\begin{figure}[H]
\centering
\begin{lstlisting}[style=scala-style, label={scala:engine-variants},
    captionpos=b, caption={The \texttt{ExtractorEngine} may be instantiated in several ways.}]
// The simplest instantiation where no actions are specified.
// Here the matches produced by a our rules are returned unmodified.
val eeNoActions = ExtractorEngine(rules)
// myActions is an object containing the implementation of any actions
// named in the rules
val eeWithActions = ExtractorEngine(rules, myActions)
// Here we specify both an actions object and a global action
val eeWithActionsAndGlobal = ExtractorEngine(rules, myActions, myGlobalAction)
// We can also choose to specify only a global action
val eeWithGlobalOnly = ExtractorEngine(rules, globalAction = myGlobalAction)
\end{lstlisting}
\end{figure}
Global actions have the same signature, but, in this context, the {\tt mentions} parameter contains all mentions found in this iteration of the engine. Any mentions produced by rule-local actions will only make it into the \texttt{State} iff they pass successfully through the global actions. By default, the global action returns its input unmodified (i.e. the input's identity).

\section{Putting it Together: the Odin API}
\label{sec:api}

In the previous sections we learned how to write token and dependency patterns using the
Odin information extraction framework. In this section, we will go
through the set up of a complete system using Odin to
extract marriage events from free text.  In Example~\ref{odin:runnable-rules}, we define a minimal grammar which we assume to be saved to the current working directory in a file named \texttt{marriage.yml}.

\begin{figure}[H]
\centering
\begin{lstlisting}[style=yaml-style, label={odin:runnable-rules},
    captionpos=b, caption={An example of a small set of rules designed to capture a marriage event and its participants. The rules that run in priority 1 make use of the output of an NER system to capture mentions for \texttt{Person}, \texttt{Location}, and \texttt{Date}.  According to the rule ``marriage-syntax-1'', a \texttt{Marriage} event requires at least one spouse and may optionally have a \texttt{Date} and \texttt{Location}.}]
- name: ner-person-or-pronouns
  label: Person
  priority: 1
  type: token
  pattern: |
   # This pattern uses the output of an NER system to
   # create a Person mention
   [entity=PERSON]+
     |
   # We will also consider some pronouns to be person Mentions
   [lemma=/^he|she|they/]

- name: ner-date
  label: [Date]
  priority: 1
  type: token
  pattern: |
    [entity=DATE]+

- name: ner-loc
  label: Location
  priority: 1
  type: token
  pattern: |
    [entity=LOCATION]+
    
# optional location and date
- name: marry-syntax-1
  label: Marriage
  priority: 2
  example: "He married Jane last June in Hawaii."
  type: dependency
  pattern: |
    # avoid negative examples by checking for "neg" relation
    trigger = [lemma=marry & !outgoing=neg]
    spouse:Person+ = (<xcomp? /^nsubj/ | dobj) conj_and?
    date:Date? = /prep_(on|in|at)/+ | tmod
    location:Location* = prep_on? /prep_(in|at)/+
\end{lstlisting}
\end{figure}

We can now use our \texttt{marriage.yml} event grammar to extract mentions from free text.  
Example~\ref{odin:runnable-engine} shows a simple Scala program to do just this.  We instantiate a \texttt{CoreNLPProcessor} which uses Stanford's CoreNLP
to parse and annotate the provided text with the attributes required by Odin
(see Table~\ref{tab:tokenattribute} for a list of the relevant attributes). 
This annotated text is stored in a \texttt{Document} which is then passed to Odin through the \texttt{EventEngine.extractFrom()} method. Finally we collect the \texttt{Marriage} mentions found by Odin and display them using the \texttt{Mention.json()} method,
which converts the mention into a \texttt{JSON} representation. A portion of this output is shown in
Example~\ref{odin:runnable-output}.

\begin{figure}[H]
\centering
\begin{lstlisting}[style=scala-style, label={odin:runnable-engine},
    captionpos=b, caption={A simple Scala program using the \texttt{marriage.yml} rules shown in Example~\ref{odin:runnable-rules}. These rules do not call any custom actions.  For an explanation of how to link rules to custom actions, please refer to Section~\ref{sec:actions}.}]
import edu.arizona.sista.odin._
import edu.arizona.sista.processors.corenlp.CoreNLPProcessor

object SimpleExample extends App {
  // two example sentences
  val text = """|John and Alice got married in Vegas last March.
                |Caesar and Cleopatra never married.
                |I think they got married.
                |Zarbon and Frederick will marry next summer.
                |She and Burt finally got married.
                |Simon and Samantha got married in Tucson on March 12, 2010 at the Desert Museum.
                |""".stripMargin

  // read rules from general-rules.yml file in resources
  val source = io.Source.fromFile("marriage.yml"))
  val rules = source.mkString
  source.close()

  // Create a simple engine without custom actions
  val extractor = ExtractorEngine(rules)

  // annotate the sentences
  val proc = new CoreNLPProcessor
  val doc = proc.annotate(text)

  // extract mentions from annotated document
  val mentions = extractor
                 .extractFrom(doc)
                 // only keep the Marriage mentions
                 .filter(_ matches "Marriage")

  // display the mentions
  mentions.foreach{ m => println(m.json(pretty=true)) }
  
}
\end{lstlisting}
\end{figure}

\begin{lstlisting}[style=scala-style, label={odin:runnable-output},
    captionpos=b, caption={An example of one of the captured \texttt{Marriage} mentions outputted as \texttt{JSON}.  The ``characterOffsets'' field corresponds to the original text provided in Example~\ref{odin:runnable-engine}).}]
{
  "type":"Event",
  "labels":["Marriage"],
  "sentence":5,
  "foundBy":"marry-syntax-1",
  "trigger":{
    "type":"TextBound",
    "tokenInterval":[4,5],
    "characterOffsets":[212,219],
    "labels":["Marriage"],
    "sentence":5,
    "foundBy":"marry-syntax-1"
  },
  "arguments":{
    "spouse":[{
      "type":"TextBound",
      "tokenInterval":[0,1],
      "characterOffsets":[189,194],
      "labels":["Person"],
      "sentence":5,
      "foundBy":"ner-person-or-pronouns"
    },{
      "type":"TextBound",
      "tokenInterval":[2,3],
      "characterOffsets":[199,207],
      "labels":["Person"],
      "sentence":5,
      "foundBy":"ner-person-or-pronouns"
    }],
    "date":[{
      "type":"TextBound",
      "tokenInterval":[8,12],
      "characterOffsets":[233,247],
      "labels":["Date"],
      "sentence":5,
      "foundBy":"ner-date"
    }],
    "location":[{
      "type":"TextBound",
      "tokenInterval":[6,7],
      "characterOffsets":[223,229],
      "labels":["Location"],
      "sentence":5,
      "foundBy":"ner-loc"
    },{
      "type":"TextBound",
      "tokenInterval":[14,16],
      "characterOffsets":[255,268],
      "labels":["Location"],
      "sentence":5,
      "foundBy":"ner-loc"
    }]
  }
}
\end{lstlisting}
An example of a complete project including details on how to specify
Odin's dependencies is available here: 

\begin{center}
\url{https://github.com/clulab/odin-examples}
\end{center}
Readers seeking a starting point for their
own projects can refer to the code in the linked repository which contains working examples covering both simple and complex scenarios.
\newpage
\bibliography{references}
\bibliographystyle{plainnat}

\end{document}